\documentclass[a4paper]{article}
\usepackage[utf8]{inputenc}
\usepackage{graphicx}
\usepackage{xcolor}
\usepackage{lineno}
\usepackage{epigraph}
\usepackage[T1]{fontenc}
\usepackage{hyperref}

\definecolor{sbase03}{HTML}{002B36}
\definecolor{sbase02}{HTML}{073642}
\definecolor{sbase01}{HTML}{93D1FA}
\definecolor{sbase00}{HTML}{657B83}
\definecolor{sbase0}{HTML}{839496}
\definecolor{sbase1}{HTML}{93A1A1}
\definecolor{sbase2}{HTML}{EEE8D5}
\definecolor{sbase3}{HTML}{FDF6E3}
\definecolor{syellow}{HTML}{B58900}
\definecolor{sorange}{HTML}{CB4B16}
\definecolor{sred}{HTML}{DC322F}
\definecolor{smagenta}{HTML}{D33682}
\definecolor{sviolet}{HTML}{6C71C4}
\definecolor{sblue}{HTML}{268BD2}
\definecolor{scyan}{HTML}{2AA198}
\definecolor{sgreen}{HTML}{859900}

\usepackage{listings}
\usepackage[a4paper, total={6in, 8in}]{geometry}
\usepackage[osf,sc]{mathpazo}
\usepackage{newtxtext,newtxmath}
\usepackage[scaled=0.90]{helvet}
\usepackage[scaled=0.85]{beramono}
\usepackage[
	backend=biber,
	citestyle=numeric, 
	bibstyle=numeric,
	maxcitenames=2]{biblatex}
\usepackage{authblk}

\title{TensorX: Extensible API for Neural Network Model Design and Deployment}
\author[1]{Davide Nunes \thanks{davex@ciencias.ulisboa.pt}}
\author[1]{Luis Antunes \thanks{xarax@ciencias.ulisboa.pt}}
\affil[1]{LASIGE, Faculdade de Ciências, Universidade de Lisboa, Portugal.}

\date{}

\begin{document}
\maketitle

\begin{abstract}
\noindent TensorX is a Python library for prototyping, design, and deployment of
complex neural network models in TensorFlow. A special emphasis is put on ease
of use, performance, and API consistency. It aims to make available high-level
components like neural network layers that are, in effect, stateful functions,
easy to compose and reuse. Its architecture allows for the expression of
patterns commonly found when building neural network models either on research
or industrial settings. Incorporating ideas from several other deep learning
libraries, it makes it easy to use components commonly found in state-of-the-art
models. The library design mixes functional dataflow computation graphs with
object-oriented neural network building blocks. TensorX combines the dynamic
nature of Python with the high-performance GPU-enabled operations of TensorFlow.

This library has minimal core dependencies (TensorFlow and NumPy) and is
distributed under \textit{Apache License 2.0} licence, encouraging its use in
both an academic and commercial settings. Full documentation, source code, and
binaries can be found in \url{https://tensorx.org/}.
    
\end{abstract}

\section{Introduction}
Machine Learning has become one of the emerging cornerstones of modern
computing. With the availability of both computational power and large amounts
of data, artificial neural networks became one of the building blocks of large
scale machine learning systems. Graphical Processing Units (GPUs), and dedicated
hardware like Tensor Processing Units (TPUs) \cite{TPUPerformance2017} reignited
the interest in large-scale vectorized computations. The performance and
architecture of such hardware makes it a perfect choice for operating on data in
vectors, matrices, and higher-dimensional arrays. This contributed to the
popularity of neural network models which, while theoretically attractive for
being universal function approximators, were mostly set aside in the past due to
their computational requirements.

Neural networks have been shown to be the state-of-the-art models in a wide
variety of tasks from text classification \cite{XLNet2020} to machine
translation \cite{EdunovTranslation2018}, or semantic image segmentation
\cite{TaoSegmentation2020}. However, replicating existing results can be
particularly challenging, not just due to computational requirements or lack of
clear experiment specifications, but because reference implementations
re-implement software components from scratch. This creates a barrier of entry
in many research tasks and makes it difficult to do iterative research. In other
words, this problem makes it difficult for the software to support the
provenance of reported results \cite{SoftwareOutput2020}. 

TensorX aims to alleviate this problem by implementing abstractions that are
usable in a wide variety of tasks to write high-level code. These components are
easy to re-use across different models and make a separation between common deep
learning technical problems and experiment definition. (e.g. re-using a
recurrent neural network cell to build complex recurrent layers, or embedding
lookup layers that handle dynamic input sequences or sparse inputs.) This
library is implemented in pure Python and it is written to be a high-level API
on top of Tensorflow \cite{TensorFlow2016}. Tensorflow is a library that allows
expressions to be defined using generalized vector data structures called
tensors or high-dimensional arrays, also the core component of the popular NumPy
library \cite{NumPy2020}. Computation graphs written with Tensorflow are
transparently transcoded to lower level machine code that can be be optimally
executed both in the CPU and GPUs along with TPUs (either in a single machine or
in a distributed cluster).

The conceptual workflow of developing and deploying neural network models is
simple: 

\begin{itemize}
	\item gather relevant data in the target domain and design a task such that
	the domain and/or the task can be captured by a model
	\item at the training phase, a learner or trainer takes the input data in
	the form of vectorial features, and outputs the state of a learned model
	\item at the inference phase, the model takes input features and outputs
	predictions or decisions, in the case of a controller
	\item at the evaluation phase, metrics are used to quantify the quality of
	the trained model
\end{itemize}

However, each of these components becomes more intricate as we need to
regularize models, evaluate training progress and model quality, decide on which
architectures to use, reuse complex modules throughout larger architectures, or
develop new components to accommodate domain knowledge or inductive bias. For
this reason, the ability to quickly prototype, standardize, and distribute
reusable components is fundamental to all scenarios, from scientific research to
applications in production in the industry. 

TensorX, similarly to e.g. Keras \cite{Keras2015}, aims to be a consistent
high-level API to build neural network models in the Python ecosystem. Keras was
absorbed into the Tensorflow codebase, departing from its initial multi-backend
design. We believe that high-level libraries should be decoupled from the
numerical computation backend. Other projects such as pytorch \cite{PyTorch2019}
also adopt this position. We believe that this makes the case for faster iterations on
bleeding-edge components, making them accessible to the community faster, while
the core backend adds operations to its code base based on scientific robustness,
and how generalizable or useful these operations are to the community.

The \footnote{\url{https://tensorx.org}}{\href{https://tensorx.org}{TensorX
website}} contains API  documentation, tutorials, and examples showcasing
multiple framework features. It also points to the public repository with the
source code, and gives instructions on how to install the library. The library
source code is distributed under the \textit{Apache License 2.0} licence.

\subsection{Related Software}
A number of high-level deep learning libraries and frameworks have emerged over
the years. This sections does not meant to present an exhaustive list of
existing libraries but rather a representation of the existing ecosystem. At
their core, most low-level libraries share the support for multi-dimensional
array transformations, automatic differentiation, and the efficient execution of
computation graphs in GPUs or similar hardware. Higher-level libraries, on the
other hand, vary in complexity of the operations supported, the number of
abstractions dedicated to neural networks in particular, and machine learning in
general, and the target domains they intend to support.

Lower-level deep learning libraries include \cite{TensorFlow2016}, PyTorch
\cite{PyTorch2019}, Chainer \cite{Chainer2019}, or \cite{MXNET2015}. More recent
additions to deep learning libraries include JAX \cite{Jax2018}, adding
automatic differentiation and GPU support to NumPy \cite{NumPy2020}, along with
graph computation optimization using the Accelerated Linear Algebra (XLA)
compiler (also used by Tensorflow). Other libraries such as DyNet
\cite{DyNet2017} offer features like dynamic batching \cite{NeubigBatching2017},
particularly useful for domains that involve the encoding of variable-length
sequences such as Natural Language Processing. Dynet ocupies somewhat a
different position in the ecosystem in that it provides both lower level
components and higher-level abstractions like Recurrent Neural Networks.

Examples of higher-level APIs and specialised frameworks include:
\cite{Fastai2020}, which is built on top of PyTorch and contains high-level
components like layer building blocks along with configurable training loops.
This perhaps the closest to TensorX in terms of scope, albeit for a different
backend library; Sonnet \cite{Sonnet2020}, with a set of high-level computation
building blocks for Tensorflow. TensorX is similar to Sonnet in the sense that
layers can be used as standalone functions, but additionally, it also includes
utilities to build and validate layer graphs, compile those graphs, and
integrate them in models that can then be trained and evaluated; TFX
\cite{TFX2017} augments Tensorflow with components for model deployment and
serving; Objax \cite{Berthelot2020}, similar to previous frameworks, but built
on top of the JAX \cite{Jax2018} backend; HuggingFace's Transformers
\cite{WolfTransformers2020}, which aims to make a specific neural network
architecture accessible to end-users with a library of pre-trained models
readily available. This is something TensorX considers for future work, but it
should be noted that the core library is instended for general purpose use.

Much like other high-level Machine Learning libraries, TensorX is built on top
of a lower level library, Tensorflow \cite{TensorFlow2016} in this case.
Tensorflow provides GPU-optimized operations, automatic differentiation, and
machine learning oriented components like optimizers. Despite libraries like
PyTorch \cite{PyTorch2019} gaining significant popularity due to its simplified
imperative programming model, when compared with previous static computation
graph definitions in TensorFlow's first version, the latest developments in the
library led to an adoption of a similar imperative computation graph definition
model. We chose to adopt Tensorflow 2 as the core due to its sizeable ecosystem,
production-oriented tools, and distributed training capabilities. TensorX
doesn't try to hide Tensorflow functionality but rather extend it and present it
in a idiomatic fashion (akin to Sonnet but with added configurable training
subroutines). Much like the Keras project \cite{Keras2015} (now integrated in
the Tensorflow codebase), we intend TensorX to be an API that simplifies neural
network rapid prototyping and deployment. We still view such high level
component libraries as something that should be developed separately as to
provide reusable state-of-the-art components without being dependent on the core
library development cycle. Also, separating the core computational components
from higher level reusable components makes the code base cleaner.

\section{TensorX Overview}
TensorX is a library designed specifically for deep learning research. It is
built on Tensorflow 2.0 \cite{TensorFlow2016}, which provides many attractive
features for neural network research. The new iteration of Tensorflow (much like
PyTorch \cite{PyTorch2019}), provides support for dynamic computation graphs
with a clear and imperative "Pythonic" syntax. At the same time, the backend
makes the benefits of optimized static computation graphs accessible through
automatic compilation of Python functions into Tensorflow graphs. TensorX takes
advantage of this and mixes an object-oriented design of stateful neural network
layers with layer graphs definitions, these in turn can be compiled into
optimized static computation graphs in a transparent fashion for the end-users.

\begin{figure*}[t!]
	\begin{center}
	\includegraphics[width=0.8\textwidth]{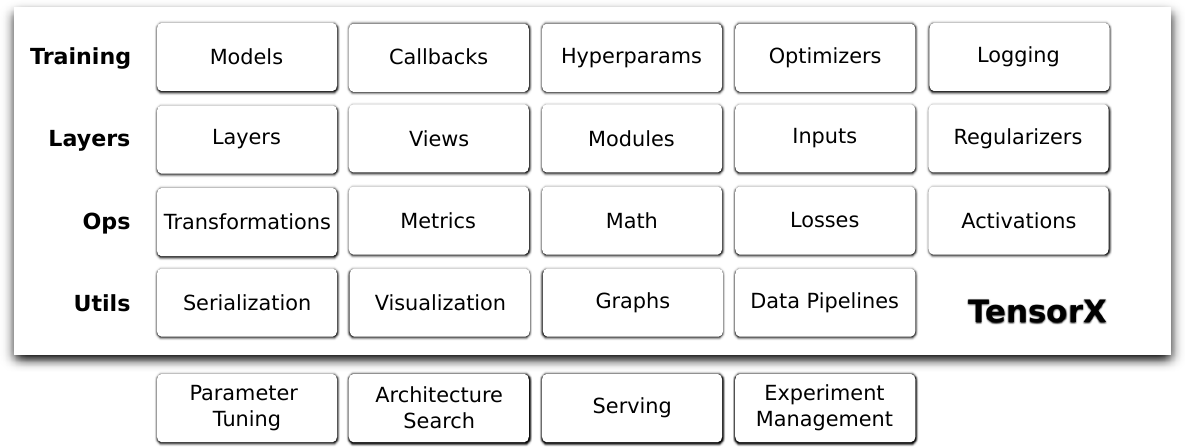}
	\caption{High-level overview of TensorX and general Machine Learning
	platform components. The other components are complementary but outside the
	scope of the TensorX package.}   
	\end{center}
	\label{fig:overview}
\end{figure*}

The main library components are illustrated in figure \ref{fig:overview}. In
this section, we will exemplify some of the features of Layer objects and layer
Graph utilities. These represent the core design decision behind the library
design and set the tone for its usability.

Hyperparameter tuning, model serving, experiment management, along with other
kind of high-level tools, while commonly found in various machine learning
toolkits, are beyond the scope of the library. The objective of TensorX is to
extend the capabilities of Tensorflow as to make research in deep neural
networks more productive both in terms of model specification and experiment
running, but the library is built with extensibility in mind so that the users
can easily contribute to it and integrate it with other tools and libraries.

\subsection{Core Components}
The core of the library is composed of \texttt{Layer} instances, layer graphs
(built automatically by \texttt{Graph} class), and the \texttt{Module} layer
which converts multiple layers into a single re-usable component that acts as
any other layer. In this section we will give a brief preview of the usage of
such components and end with a summary of how these components interact with
each other.


\paragraph{Layer} At the core neural network building blocks in the TensorX
library are Layer objects. Semantically speaking, a layer is an object that can
have multiple inputs, an inner state, and a computation function that is applied
to its inputs (and depends on the current inner state). Each layer has a single
output. In essence, we can say that a \texttt{Layer} instance is a stateful
function. 

Layer subclasses can range from simple linear transformations (e.g. in the form
$y = Wx + b$ where $W$ is a weight matrix and $b$ a vector with biases) to more
complex structures used to build recurrent neural networks such as Long~
short-term~memory~(LSTM) layers \cite{HochreiterLSTM1997} or attention
mechanisms \cite{BahdanauAttention2014}, or even layers used for regularization
such as \texttt{Dropout} \cite{SrivastavaDropout2014}.

\begin{figure}[h!]
\noindent\begin{minipage}{.5\textwidth}
\vspace{0.1cm}
\begin{lstlisting}
def init_state(self):
	state = super().init_state()
	x = ... 
	h = ...
	w = Linear(x, self.n_units, ...)
	u = Linear(h, self.n_units, ...)

	add_wu = Add(w, u)
	output = Activation(add_wu, tx.tanh)
						
	state.rnn_cell = Module([x, h],output)
	return state

def compute(self, x, *h):
	return self.rnn_cell.compute(x,*h)

\end{lstlisting}
\end{minipage}
\begin{minipage}{.48\textwidth}
\begin{center}
	\includegraphics[width=1\linewidth]{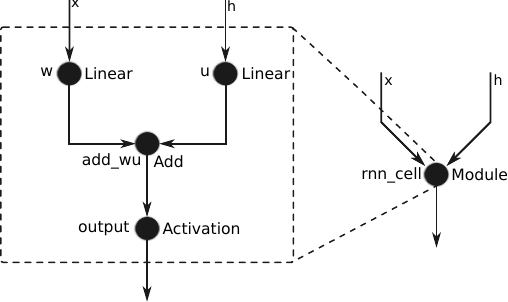}
\end{center}	
\end{minipage}
\caption{Example of recurrent neural network (RNN) cell definition using
\texttt{Module} to consolidate a layer graph into a single component that
can later be executed.}
\label{example:reuse}
\end{figure}

Figure \ref{example:reuse} shows an example of basic layer used to construct a
computation graph with multiple layers. We can also see how to reuse existing
layers in such a way that their internal state is shared between layer
instances.

\begin{figure}[h]
\noindent\begin{minipage}{.5\textwidth}
\vspace{0.2cm}
\begin{lstlisting}
import tensorflow as tf
import tensorx as tx

# stateful input placeholder
x1 = tx.Input(n_units=2)
x1.value = tf.random.uniform([2, 2])

# y = Wx + b
l1 = tx.Linear(x1, n_units=3)
a1 = tx.Activation(l1, tx.relu)
l2 = tx.Linear(a1,n_units=4)

d1 = tx.Dropout(a1,probability=0.4)
l3 = l2.reuse_with(d1)
\end{lstlisting}
\end{minipage}
\begin{minipage}{.48\textwidth}
\includegraphics[width=0.8\linewidth]{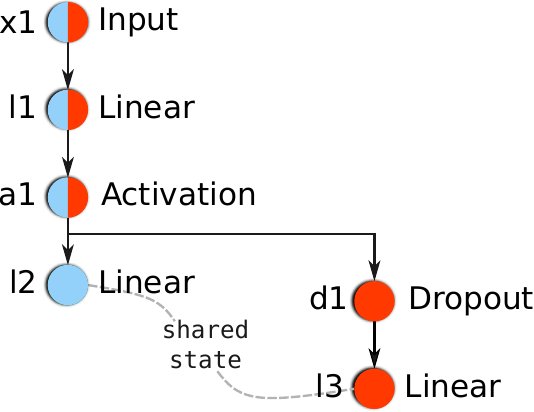}
\end{minipage}
\caption{Example with basic layer creation and state re-use. Introducing a
		 regularization layer (Dropout) between layers in another pre-existing
		 graph. Each output is at the same time a layer object and the end-node
		 of a computation graph.}
\label{example:module}
\end{figure}

A \texttt{Layer} object is simultaneously a stateful function and the end-node
of a computation graph. Executing a layer will execute the entire graph ending
in that specific node. If we only want to execute a layer computation on a given
set of inputs, we can use the \texttt{compute(*inputs)} method. Note also that
\texttt{Input} is a special layer that has no inputs, instead, this is used as a
stateful placeholder that stores the inputs for the current computation graph.

\paragraph{Module}

A \texttt{Module} is a special utility layer that transforms a computation graph
between into a new \texttt{Layer} object. The Module class traces a graph
between the given output and its inputs, determine if the graph is valid, and
transforms the layers into a single layer/stateful function. A use case for this
feature is the development of new TensorX layers, as it allows us to use the
state initialization procedure to define complex layer graphs, these can then be
transformed into a single module that is executed by the
\texttt{(compute(*inputs)} method. Figure \ref{example:module} shows an example
of \texttt{Module} being used to create a recurrent neural network (RNN) cell.

\paragraph{Graph} In TensorX, as we have seen previously, by connecting multiple
layers to each other, we build layer graphs. These are in effect directed
acyclic graphs (DAG) defining a given computation over inputs. To aid with
validation and execution of neural network layer graphs, TensorX has a
\texttt{Graph} utility class. The \texttt{Graph} class allows for automatic
graph construction from output nodes (by recursively visiting each node's
inputs). It also facilitates transversal by dependency ordering along with
conversion of arbitrary graphs to functions. Moreover, this conversion allows
for TensorX graphs to be compiled in to Tensorflow static computation graphs.

We take advantage of Tensorflow's graph optimization system to optimize
layer graph computations. This system improves the performance of TensorFlow
computations through computation graph simplifications and other high-level optimizations.
By converting layers into functions that are then trace-compiled into an
optimized static graph, we get the best of both worlds (\texttt{Layer} instances
are easy to debug in eager mode, and layer graphs are transparently compiled
into optimized Tensorflow graphs).

\begin{figure}[h!]
\noindent\begin{minipage}{.49\textwidth}
\vspace{0.2cm}
\begin{lstlisting}
x1 = Input(n_units=2)
x2 = Input(n_units=4)
l1 = Linear(x1,4)
l2 = Add(l1,x2)
l3 = Linear(l2,2)

g = Graph.build(outputs=l3,
							inputs=[x1,x2])

fn = g.as_function(compile=True)
\end{lstlisting}
\end{minipage}
\begin{minipage}{.49\textwidth}
\begin{lstlisting}
# fn is holding the following function
# tf.function decorator added if compile is True
@tf.function 
def compiled_graph():
	x1 = layers["x1"].compute()
	x2 = layers["x2"].compute()
	l1 = layers["l1"].compute(x1)
	l2 = layers["l2"].compute(l1,x2)
	l3 = layers["l3"].compute(l2)
	return l3
\end{lstlisting}
\end{minipage}
\caption{Automatic graph building example. The graph \texttt{g} is traced from
the output nodes until the inputs are reached. Graphs are also capable of being
converted into functions as demonstrated. TensorX uses the dynamic nature of
python to create a new function object that can then be traced-compiled by
Tensorflow into an optimized static computation graph.}
\label{example:graph}
\end{figure}

\begin{figure}[th]
	\begin{center}
	\includegraphics[width=0.85\textwidth]{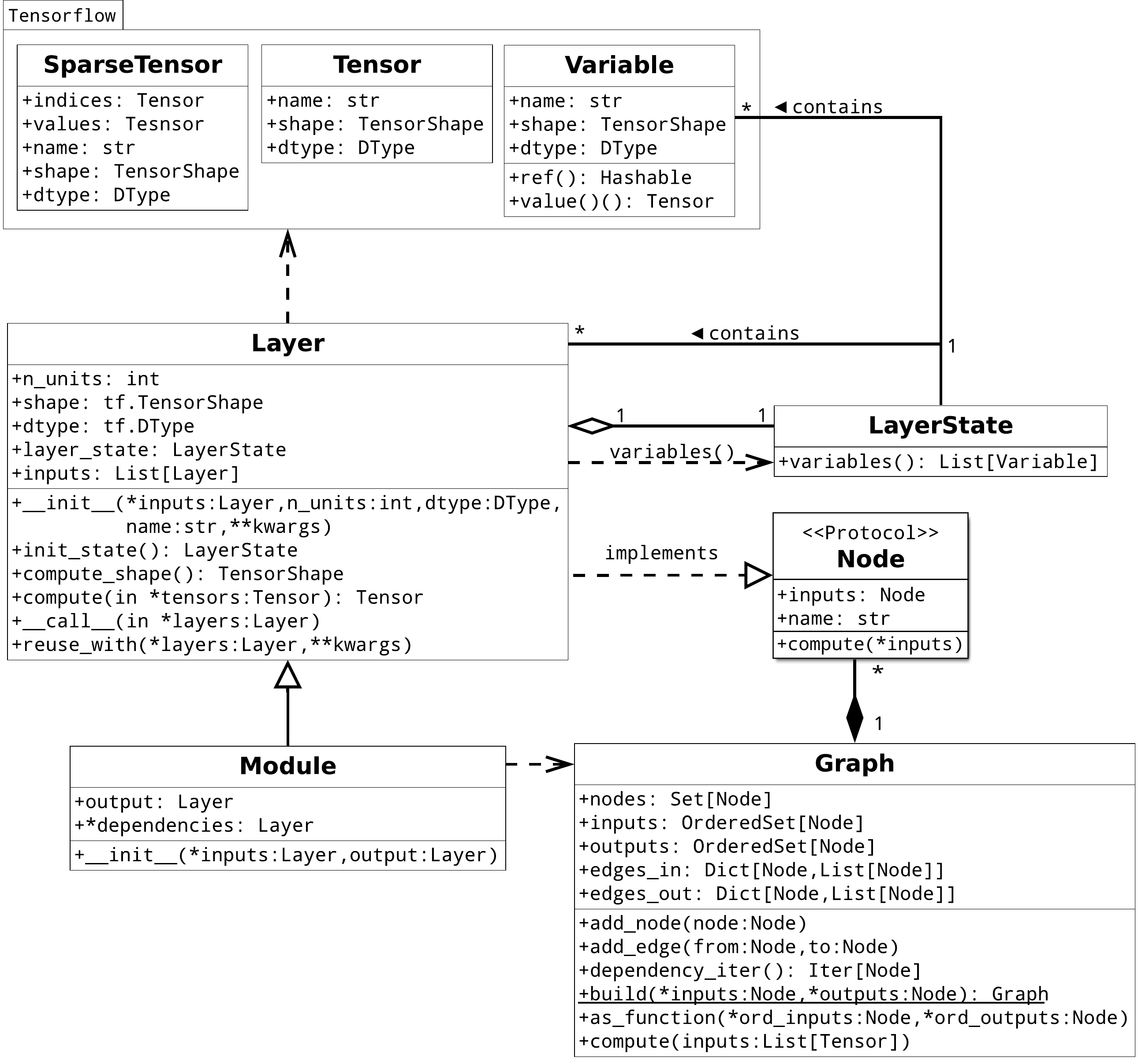}
	\caption{Summarised UML diagram for the main building blocks in the TensorX library.}   
	\end{center}
	\label{fig:layer_uml}
\end{figure}

Figure \ref{fig:layer_uml} shows a summary UML diagram of the previously
mentioned components, along with their basic interaction. While there are many
ready to use layers in the library, from different types of recurrent neural
network cells to sequence lookup, convolution layers among others, this short
excerpt illustrates the main design decisions behind the library and set the
tone for the usability of the API TensorX provides.

As we can see, layers have access to basic Tensorflow constructs like
\texttt{Tensor}, \texttt{SparseTensor}, or \texttt{Variable}, and encapsulate
the stateful computations each basic layer provides. Layer states are decoupled
from layers as to avoid the need for referencing each layer sharing a given
state to propagate a modified member. A \texttt{Module}, as previously
discussed, is a special layer that makes use of the \texttt{Graph} utility to
encapsulate complex layer graphs as a single reusable object. The graph utility
itself is a general data structure that uses only inputs as a transversal
method, and a \texttt{compute} method to convert a graph into python function.

For more documentation and examples refer to the library documentation website.
The previous are the basic TensorX building blocks used to construct most of the
other components (e.g. the \texttt{training} module contains training utilities
that make use of \texttt{Graph} instances to encapsulate a model
\textit{inference}, \textit{training}, and \textit{evaluation} graphs).

\section{Conclusion and Future Work}
Deep neural networks continue to play a major role in fields like Natural
Language Processing, Computer Vision, Reinforcement Learning, and Machine
Learning in general. As these models and methodology continue to gain traction
and technology transfer makes them especially attractive to build real-world
applications, it is important for model building, and experiment deployment
tools to be accessible both in an research, industrial context to end-users.
TensorX aims to be an open-source library that fulfils that role allows the
community to built upon this work and contribute with relevant components
--making state-of-the-art advancements widely available to everyone without
depending on the core development cycle of its backend library Tensorflow.

Future work includes making a set of full models like Transformers readily
available using components from the library, full integration with distributed
training from Tensorflow and actor-based distributed computing frameworks such
as Ray \cite{MoritzRay2018}. Finally, our goal is to integrate TensorX with
other experiment tracking and monitoring platforms, extending the existing tools
to a wider community. TensorX aims to do a couple of things well rather than
encapsulating all the possible needs of the Machine Learning community under a
single library, as such the goal is to maintain an extendable open platform with
solid foundations from which end-users can build.

\printbibliography

\end{document}